\documentclass[a4paper, 10pt, conference]{ieeeconf}      

\IEEEoverridecommandlockouts                              

\usepackage{times}
\usepackage{epsfig}
\usepackage{graphicx}
\usepackage{algorithm}
\usepackage{algpseudocode}
\usepackage{subcaption}
\usepackage{array}
\usepackage{amsmath}
\usepackage{amssymb}
\usepackage{balance}
\usepackage{hyperref}
\usepackage{booktabs}
\newcolumntype{P}[1]{>{\centering\arraybackslash}p{#1}}

\usepackage[flushleft]{threeparttable}
\usepackage{xcolor}

\usepackage[T1]{fontenc} 

\title{\LARGE \bf
Vision-Guided Active Tactile Perception for Crack Detection and Reconstruction}

\author{Jiaqi Jiang, Guanqun Cao, Daniel Fernandes Gomes and Shan Luo  
\thanks{All the authors are with the Department of Computer Science, University of Liverpool, Liverpool L69 3BX, U.K. Emails:\{jiaqi.jiang, g.cao, danfergo, shan.luo\}@liverpool.ac.uk.}%
}

\begin{document}

\maketitle
\thispagestyle{empty}
\pagestyle{empty}

\begin{abstract}
Crack detection is of great significance for monitoring the integrity and well-being of the infrastructure such as bridges and underground pipelines, which are harsh environments for people to access. 
In recent years, computer vision techniques have been applied in detecting cracks in concrete structures. However, they suffer from variances in light conditions and shadows, lacking robustness and resulting in many false positives. To address the uncertainty in vision, human inspectors actively touch the surface of the structures, guided by vision, which has not been explored in autonomous crack detection. 
In this paper, we propose a novel approach to detect and reconstruct cracks in concrete structures using vision-guided active tactile perception.
Given an RGB-D image of a structure, the rough profile of the crack in the structure surface will first be segmented with a fine-tuned Deep Convolutional Neural Networks, and a set of contact points are generated to guide the collection of tactile images by a camera-based optical tactile sensor. When contacts are made, a pixel-wise mask of the crack can be obtained from the tactile images and therefore the profile of the crack can be refined by aligning the RGB-D image and the tactile images.
Extensive experiment results have shown that the proposed method improves the effectiveness and robustness of crack detection and reconstruction significantly, compared to crack detection with vision only, and has the potential to enable robots to help humans with the inspection and repair of the concrete infrastructure.



\end{abstract}

\section{INTRODUCTION}
Cracks in the infrastructures surface are important indicators for assessing the condition of buildings and need to be repaired timely for preventing the expansion of potential risks. However, manual crack detection and maintenance are not only time-consuming and expensive, but also pose health risks to human workers in harsh and complex environments such as dams and underground pipelines. Hence, development of an effective and robust crack detection system will be significant for the substitution of inspection workers and the development of smart cities \cite{du2018sensable}.

During the past two decades, the problem of crack detection attracts wide attention from researchers and has been widely explored and developed \cite{zakeri2017image}. However, most of the previous works address this task based on only one single modality such as acoustic data \cite{chakraborty2019early} \cite{louhi2020acoustic}, laser-scanned image \cite{zhou2020deep}, optical fiber signal \cite{palermo2020implementing}, and RGB images \cite{fang2020novel}. In this case, those methods using the high-precision modality such as acoustic data cannot perceive objects quickly due to the limitation of the small receptive field. Moreover, they could not correct the detection results with additional information if a crack is not detected or a noisy pattern is falsely detected as a crack.


When humans intend to perceive objects, they gather information and draw a solid conclusion with their eyes, ears and hands together. 
In crack detection tasks, skilled inspectors usually first look at the surface to find areas with similar color or shape characteristics to cracks, and then use their fingers or specific tools such as a hammer or ultrasonic device to further inspect those areas instead of traversing all regions. In this way, they can fully utilize the properties of different modalities including the efficiency of visual sense with a large receptive field and the accurateness of tactile sense or acoustic sense.

Inspired by those observations, we propose a novel vision-guided active tactile perception approach for crack detection and reconstruction in this paper. This method integrates visual and tactile senses so that they aid each other and improve the perception on cracks. On the one hand, visual sense with a large receptive field can achieve a quick search for finding the candidate regions containing cracks, which can reduce the area of tactile perception and improve its effectiveness. On the other hand, tactile sense can enhance the robustness of crack detection thanks to the characteristics of being less susceptible to light and noise, and can reconstruct the crack shape precisely with high-resolution tactile images obtained by a camera based optical tactile sensor. 

To evaluate our proposed method, we collected a visual dataset and a tactile dataset with 10 mock-up structures that were manufactured using 3D printing. Half of those objects only contain real cracks, and the rest contain both real cracks and fake cracks that are paintings on the surface. The extensive experiment results demonstrate that our approach is capable of distinguishing real cracks from fake cracks, and reconstructing the crack shape quickly and accurately. Compared to the visual method, our approach 
achieves a significant reduction of mean distance error from 0.82mm to 0.24mm. Furthermore, the proposed method is more than 10 times faster than passive tactile perception in terms of tactile data collection time.

Our contributions can be summarised as follows:
\begin{itemize}
\item A novel vision-guided active tactile perception framework is proposed, for the first time, for crack detection and reconstruction;
\item A touch point generation method is designed for the vision-guided active tactile perception;
\item A tactile crack perception method is developed that can suppress the false positives of visual detection and reconstruct the crack with a millimeter-level resolution.
\end{itemize}



\section{Related Works}
\subsection{Vision-Based Crack Detection}
Recently the study of image-based crack detection has drawn increasing attention from researchers in the fields of both Computer Vision and Civil Engineering. Early vision-based cracks detection methods mainly rely on conventional image processing methods such as edge detection \cite{prasanna2014automated} \cite{hanzaei2017automatic} and thresholding \cite{salari2011pavement} \cite{zhang2017efficient}. However, those methods are susceptible to light changes and scanning noises due to the subjective parameter selections for pre-processing and edge detection. Moreover, there are also works using conventional machine learning approaches such as Support Vector Machine \cite{bu2015crack} \cite{chen2017texture}, Ada-Boost \cite{zalama2014road} and Na\"ive Bayes classifier \cite{schmugge2014automatic}. Nonetheless, the features used in these traditional machine learning methods such as LBP, wavelet and histogram can only represent one or two layers of abstraction, which are unable to capture features of cracks in surfaces with complex backgrounds.

In the recent years, works using deep neural networks for crack detection have emerged \cite{chen2017nb,pauly2017deeper}. Based on the fineness of detection results, those deep learning-based methods can be summarised as two types of approaches: box-level detection and pixel-wise detection. Typical box-level crack detection approaches firstly generate a set of rectangular frames to locate the cracks using patch scanning \cite{chen2017nb} or region-based CNN networks \cite{fang2020novel} and then filter out false positives with a Na\"ive Bayes classifier \cite{chen2017nb} or a Bayesian integration algorithm \cite{fang2020novel}. However, the rectangle shape results are not precise and are limited with regard to complex crack distribution and irregular infrastructure surface. The more recent development in deep learning-based defects perception is pixel-wise crack detection. In \cite{zou2018deepcrack}, a novel encoder-decoder network is proposed, which uses hierarchical convolutional feature learning to separate the cracks from background. In \cite{zhang2020crackgan}, a deep generative adversarial network is proposed for pavement crack detection, which is trained end-to-end with a crack-patch-only supervised method to overcome the local minimum problem. 


\subsection{Active Tactile Perception}

Tactile sensing plays an important role in robot perception and has been applied to a number of different tasks such as object recognition \cite{luo2017robotic}, material property analysis \cite{luo2018vitac}, and shape exploration \cite{lepora2019pixels}. 
Most early studies related to tactile perception are based on passive data acquisition processes and are considered simply as a forward process \cite{luo2018vitac,lee2019touching}. Nonetheless, humans usually plan their actions with a goal and actively refine their sensations (especially the tactile sensation) rather than passively gather information. Inspired by this, there has been growing interest in active tactile perception \cite{lepora2017exploratory,martinez2017feeling}. 
In \cite{yuan2018active}, an active tactile perception method using pre-specified control laws was proposed for clothing material recognition.

However, there has been no works on crack detection with active tactile perception yet. In \cite{palermo2020implementing}, tactile sensing was explored for crack detection and characterization, but in this work tactile data was collected passively and the position of the crack cannot be localized accurately. Our proposed vision-guided active tactile perception is the first work that incorporates visual information into active tactile perception for crack detection and reconstruction. It can actively perceive the crack region with the guidance of vision, and reconstruct crack shape precisely with a high-resolution camera-based tactile sensor.






\section{Methodology} 
\begin{figure*}[h]
\begin{center}
   \includegraphics[width=\linewidth]{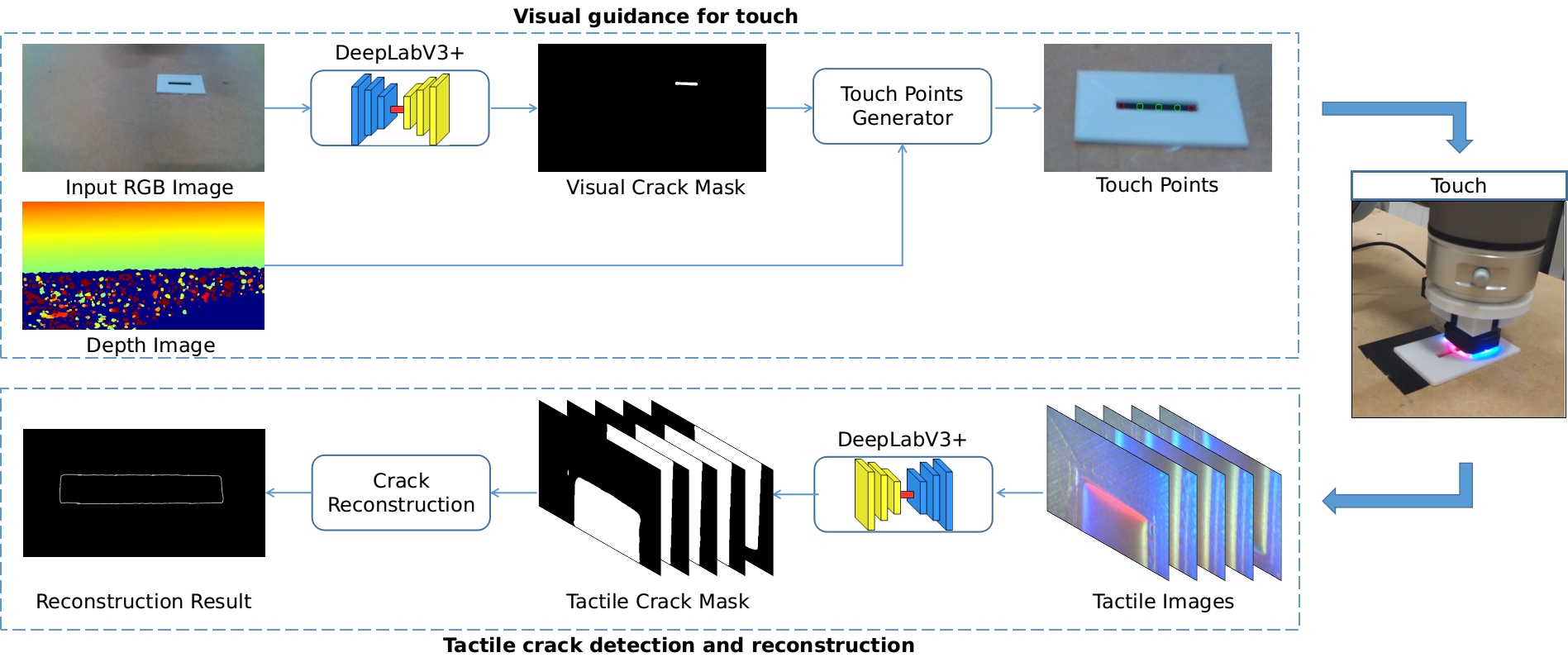}
\end{center}
   \caption{An overview of our vision-guided active tactile crack detection and reconstruction method. \textbf{Top row (from left to right):} The Deeplabv3+ model is used to segment the cracks in the visual image. Given the visual crack mask and the depth image, a set of contact points are generated to guide the collection of tactile images. \textbf{Bottom row (from right to left):} Another deep convolutional network is used to segment the crack in the collected tactile images. Given the detected tactile crack mask, the crack shape are reconstructed based on the geometrical model of the GelSight sensor and the coordinate transforming relation between the tactile  sensor coordinate and the world coordinate. }
\label{fig:overview}
\end{figure*}

In this work, we propose a vision-guided active tactile perception framework that can detect and reconstruct cracks efficiently and precisely, with an overview of the framework illustrated in Fig. \ref{fig:overview}. Given a single RGB-D image of cracks in a structure, our method first uses the color image as input to a deep convolutional network to predict a mask of the cracks. Then the predicted mask and the depth image are used to generate a set of contact points, which directs the collection of the tactile images with a camera-based tactile sensor. Another deep convolutional network is applied to predict the masks of cracks in the collected tactile images that are used to refine the visual detection result and reconstruct an accurate cracks profile.




\subsection{Visual Guidance for Touch}\label{section:IIIA}
A deep semantic segmentation network is first utilized to predict pixel-wise masks of cracks. To guide the touch sensing, the contact points with cracks skeletons are then generated. Details of these two steps are given below.


\textbf{Visual Crack Segmentation.} 
The visual crack segmentation is treated as a semantic segmentation problem that predicts each pixel of the input image into one of two semantic classes: (a) background (b) cracks. To this end, we use the Deeplabv3+ model \cite{deeplabv3plus2018} to segment the cracks in visual images that is a state-of-art deep learning model for semantic image segmentation. A ResNet-101 backbone \cite{he2016deep} is used. Since the number of pixels of the background is much larger than that of the cracks, the network may easily converge to the status that treats all the pixels as background. To address this issue, we use the original images as input instead of resizing it to a smaller size as done in the previous works \cite{pauly2017deeper} and use a weighted cross-entropy loss with crack pixels weighted 10x more than background pixels. Moreover, we set the output stride value as 8 as smaller values give finer details in the output mask. In the training process, we start with a model pre-trained for semantic segmentation on the COCO dataset \cite{lin2014microsoft} and use the following hyperparameters: SGD Optimizer with constant learning late of 1e-6, momentum 0.9 and weight decay 5e-4. 





\textbf{Contact Points Generation.}
Given the predicted pixel-wise crack mask in the color image, we can extract the skeleton of each crack mask with pattern thinning method \cite{zhang1984fast}. We define two types of keypoints (i.e., end points and branch points) and minimal edges that represent the topology of the crack pattern: 
\begin{itemize}
    \item  End points: if they have less than two neighbors.  
    \item  Branch points: if they have more than two neighbors. 
    \item  Minimal edge $E_{ij}$: if there is a continuous path between two keypoints $p_i$ and $p_j$ and all points on the path are neither end points nor branch points. 
\end{itemize}

For every minimal edge $E_{ij}$ which consists of a number of ordered points, the keypoint $p_i$ is initially selected as the current contact point $p_{current}$. Then we iteratively choose the next contact point $p_k$ using the following formula: 
\begin{equation}
\begin{array}{l}
\max \limits_{k} D[p_{current},p_k] \\
\text { s.t. } D[p_{current},p_k]<d 
\end{array}
\end{equation}
where $D(p_{current},p_k)$ is the distance between two points in world frame. The hyper-parameter $d$ is the threshold of the distance between two points that is related to the coverage and speed of tactile perception. A smaller $d$ will increase the coverage while reducing the perception speed. In our case, $d$ is empirically set to four fifths of the tactile sensor's view length. As shown in Fig. \ref{fig:overview}, the end-point pixels and the generated contact points are tagged with red dots and green dots, respectively. For each contact points $p_i$, the yaw angle of the end-effector is parallel to the vector $<p_i,p_n>$, where $p_n$ is the nearest contact point to $p_i$, so that the end-effector can contact the surface perpendicularly.

\subsection{Active Tactile Crack Perception}

\textbf{Tactile Crack Detection.}
To address the problem of false positives in the visual crack detection due to light changes and shadows, we apply tactile information to refine the vision-based detection results and reconstruct cracks shape in 3D space. Firstly, we control the robotic arm equipped with a camera based Gelsight tactile sensor \cite{yuan2018active} to collect tactile images autonomously at the generated contact points in Section~\ref{section:IIIA}.

\begin{figure}[h]
\begin{center}
   \includegraphics[width=0.6\linewidth]{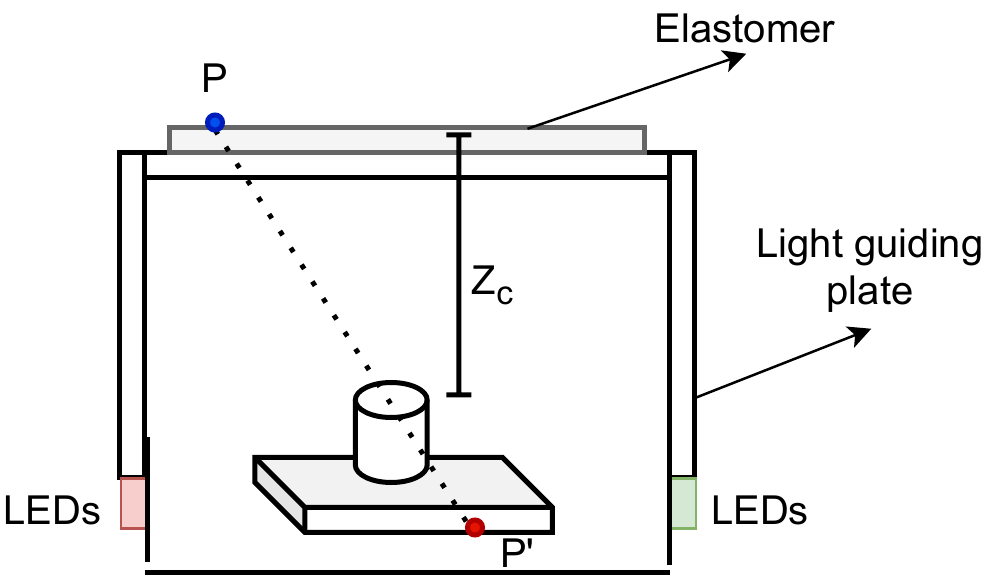}
\end{center}
   \caption{The geometrical model of the GelSight sensor. The webcam at the bottom captures the deformations of the elastomer and the LEDs project light to illuminate the space. }
\label{fig:model}
\end{figure}

The GelSight sensor is a camera-based optical tactile sensor that can capture fine details of the object surface. As shown in Fig.~\ref{fig:model}, a webcam\footnote{We use ``webcam'' to refer to the camera used in the optical tactile sensor to differ it from the visual camera used for the vision modality.} is placed under an elastomer and captures the deformations of the elastomer when it interacts with the object. The sensor has a flat surface and a view range of $14 mm \times 10.5 mm$ and can capture tactile images at a frequency of 30 Hz \cite{cao2020spatio}.


The collected tactile images are fed to another deep convolutional network (Deeplabv3+  \cite{deeplabv3plus2018} with ResNet-101  \cite{he2016deep}). Since the number of background pixels is similar to that of the background pixels in tactile images, we use the vanilla cross-entropy loss instead of the weighted cross-entropy loss in Section~\ref{section:IIIA}.
Using the predicted masks in tactile images, we can double check the visual segmentation results. Any minimal edge will be set as false positives and deleted from visual segmentation results, if there are more than two tactile images whose predicted crack area is smaller than a predefined threshold. The threshold is empirically set as one fiftieth of the total number of pixels in the tactile image.

\textbf{Tactile Crack Reconstruction.}
In the task of crack perception, it is crucial to obtain the shape and size of the crack when assessing its potential risk to the building and infrastructure. However, limited by the accuracy of the depth camera, current vision-based reconstruction methods are not able to reconstruct small cracks precisely. To this end, we use tactile images obtained from the GelSight sensor to reconstruct the cracks, whose spatial resolution is about 20 to 30 microns. 


First, we predict the location of cracks on the surface of the GelSight sensor, given the detected boundaries of pixel-wise masks in the tactile images. To simplify the problem, we model the webcam using a pinhole camera model, and treat the surface of GelSight sensor as a flat plane that is perpendicular to the webcam's $z$ axis. Hence, the transformation between a contact point $P=[X_{c}, Y_{c}, Z_{c}]^{T}$ in the tactile sensor coordinates (the tactile sensor take the centre of the webcam as the origin) to the pixel $P'=[u, v]^{T}$ in the tactile image coordinates can be calculated:
\begin{equation}Z_{c}
\left[\begin{array}{l}
u \\
v \\
1
\end{array}\right]=K
\left[\begin{array}{c}
X_{c} \\
Y_{c} \\
Z_{c} \\
1
\end{array}\right]
\end{equation}
where $Z_c$ is the distance from the optical center to the elastomer surface of the tactile sensor. $K$ is the matrix of the intrinsic parameters of the webcam and can be represented as:
\begin{equation}
K=\left[\begin{array}{llll}
\frac {f}{d_{x}} & 0 & u_{0} & 0 \\
0 & \frac {f}{d_{y}} & v_{0} & 0 \\
0 & 0 & 1 & 0
\end{array}\right]
\end{equation}
where $f$ is the focal length of the webcam, $d_{x}$ and $d_{y}$ denote the pixel size, and $(u_{0}, v_{0})$ is the center point of the tactile image.

After obtaining the position $P$ in the tactile sensor coordinates, we can calculate its position $P_W$ in the world frame:
\begin{equation}
P_{W}=T_{E}^{W} T_{C}^{E} P
\end{equation}
where $T_{C}^{E}$ and $T_{E}^{W}$ are the transformation matrix from the tactile sensor coordinates to the end-effector coordinate system, and from the end-effector coordinate system to the world coordinate system, respectively.

\section{Experiment Setup}
\begin{figure}[h]
\begin{center}
   \includegraphics[width=1\linewidth]{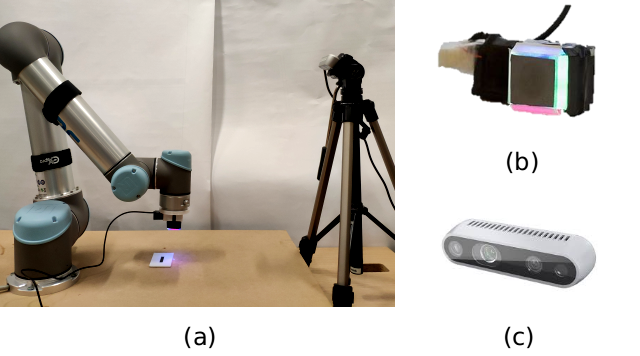}
\end{center}
   \caption{ a) The UR5 robot setup with GelSight sensor and Realsense D435i; b) GelSight sensor; c) Realsense D435i.}
\label{fig:robotsetup}
\end{figure}

In this section, we introduce the robot setup used for data collection and experiments. An overview of our robot setup is shown in Fig. \ref{fig:robotsetup}.


\subsection{3D Printed Structures with Cracks }
\begin{figure}[h]
\begin{center}
   \includegraphics[width=\linewidth]{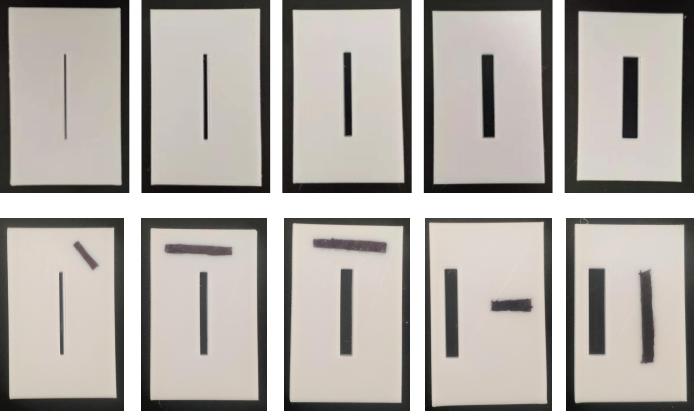}
\end{center}
   \caption{Sample structures used for collecting the visual and tactile dataset. \textbf{Top row}: printed structures with real cracks (holes). \textbf{Bottom row}: printed pad with fake cracks (painted black blocks).}
\label{fig:samples}
\end{figure}

In this work, we follow part of the data acquisition protocol in \cite{palermo2020implementing}.
A set of 10 structures with cracks of different widths (holes in the structures) were manufactured with PLA plastic using an i3 Mega 3D printer from ANYCUBIC. To test the robustness of the proposed method, we also paint several fake cracks on the surfaces of the structures. The samples are shown in Fig. \ref{fig:samples}. 


\subsection{Visual Data Collection}
We put those 3D-printed structures on a table and took 10 images of each structure with real cracks that are used for training the visual model. Then we take 3 images for each object separately for the test. All images were taken with a RealSense D435i camera.

\subsection{Tactile Data Collection}
\begin{figure}[h]
\begin{center}
   \includegraphics[width=\linewidth]{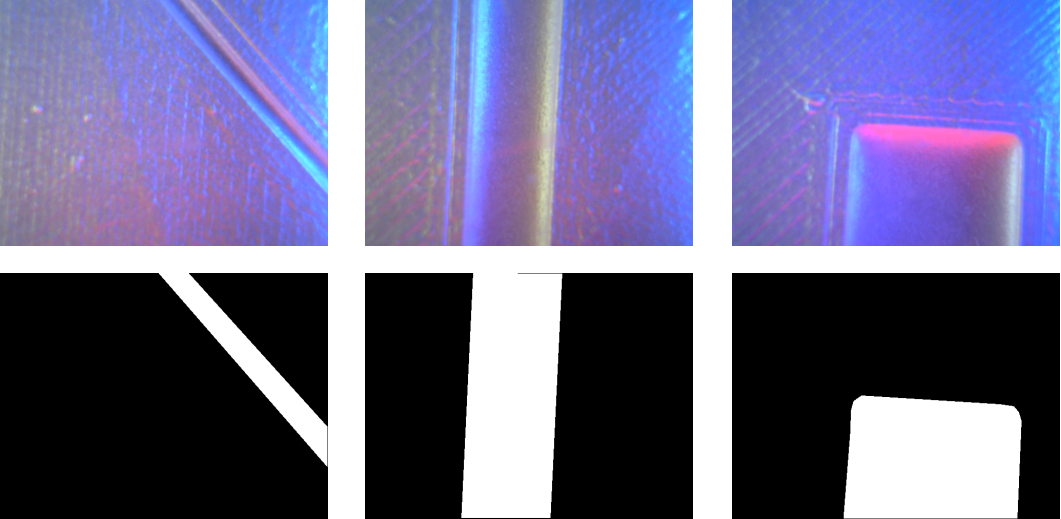}
\end{center}
   \caption{Visualization of the tactile images and their annotations for different cracks. }
\label{fig:tactileimages}
\end{figure}

The setup of tactile data collection is composed of two parts: a 6-DOF UR5 collaborative robot arm from Universal Robots and a GelSight sensor mounted on a 3D-printed end-effector. To collect the tactile data autonomously and repeatedly, we build a data collection software using Robot Operating System (ROS). The software can control the robot arm to move across the surface of a structure following the pre-defined initial position, steps, and step length in x-axis and y-axis. As for each position, we rotate the sensor about the axis perpendicular to the surface with different angles so that the dataset can be more generalized. As for each contact, the robot arm will stop moving and the GelSight records one tactile image when the pressure reaches a threshold. In this way, we can capture tactile data of good quality and also avoid the unnecessary protective stops of the robot arm. In total, 544 valid tactile images were collected and split into training and test sets (370 and 174 for each, respectively), with some samples and their annotations shown in Fig. \ref{fig:tactileimages}.  

\section{Experiment Results}

\subsection{Experiments on Network Architectures}
\begin{figure}[h]
\begin{center}
   \includegraphics[width=\linewidth]{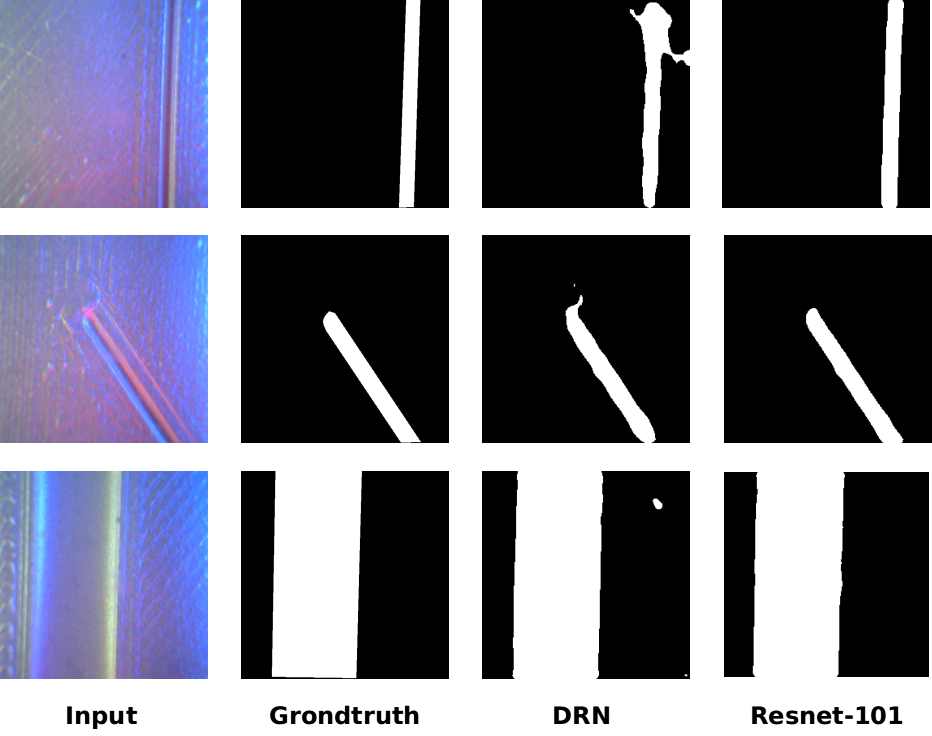}
\end{center}
   \caption{Visual comparison of different backbones. }
\label{fig:visual comparison}
\end{figure}

To evaluate the accuracy of the tactile crack segmentation, we use both Intersection over Union (IoU) and True Positive rate (TP). The results of different models and input image sizes are listed in Table \ref{tab:re1}. Two different backbones was used for Deeplabv3+: ResNet-101 and DRN-54 (Dilated Residual Network). Example segmentation results of ResNet-101 and DRN-54 with an input size of 256$\times$256 are shown in Fig. \ref{fig:visual comparison}. The results indicate that compared to Resnet-101, DRN introduces more noise over training and perform worse for the tactile crack segmentation due to the fact that it has a larger receptive field. 

We also experimented with different input image sizes. The results show that a smaller input image size leads to better performance, which is surprising as images of a higher resolution may contain more details. One possible reason is that a certain amount of noise is generated when resizing the width of the tactile image from 480 to 512.

\begin{table}[h]
	\centering
		\caption{Network Architecture For Tactile Crack Segmentation.}
		\label{tab:re1}
        \scalebox{1}{
		\begin{tabular}{c| c | c | c }
			\hline
			Backbone & Input Size & IoU & TP    \\
			\hline
			\hline
    		DRN     & 512 & 0.93 & 95.48 \\
    		DRN     & 256 & 0.93 & 97.32 \\
    		ResNet-101      & 512 & 0.94 & 96.65\\
    		ResNet-101[ours]   & 256 & \textbf{0.97} & \textbf{98.87}\\
    		
    		\hline
		\end{tabular}}
\end{table}



\subsection{Crack Detection Results} 
Similar to the evaluation method used in network architectures experiments, we evaluate our proposed methods based on standard evaluation metrics of pixel accuracy (pixAcc) and IoU. The segmentation results of using only visual information and using both visual and tactile information are summarized in Table \ref{tab:re2}. The results show that both the pixel accuracy and IoU of visual semantic segmentation have a significant drop due to the existence of fake painting cracks from 0.899 to 0.866 and from 0.504 to 0.376, respectively. It is also shown that after using tactile information to find the fake paintings and refine the visual detection results, the performance of cracks detection has been improved effectively. Due to the fact that cracks on the RGB images only take up fewer pixels compared to background, the pixel accuracy does not change dramatically as IoU.



%
\begin{table}[h]
	\centering
		\caption{Crack Detection Accuracy.}
		\label{tab:re2}
        \scalebox{1}{
		\begin{tabular}{c| c | c | c }
			\hline
			Modalities & Fake Painting & pixAcc & IoU  \\
			\hline
			\hline
    		vision     &  $\times$ & 0.899 & 0.504\\
    		vision     & \checkmark & 0.866 & 0.376\\
    		vision-tactile  & \checkmark & \textbf{0.909} & \textbf{0.636}\\
    		\hline
		\end{tabular}}
\end{table}

\subsection{Reconstruction Results} 
As for the crack reconstruction task, we use the mean, max and standard deviation (SD) of the shortest distance between the actual crack shape and the reconstructed crack location to evaluate the accuracy of our proposed method. There are four methods used for comparison. The vision method uses point clouds recovered through visual detection and depth information to represent cracks. In order to reduce the impact of depth information accuracy on reconstruction, the aligned vision method projects the point cloud to the table surface. The passive tactile method collects the tactile images through traversing the whole 3D printed structure surface. 

\begin{figure}[h]
\begin{center}
   \includegraphics[width=\linewidth]{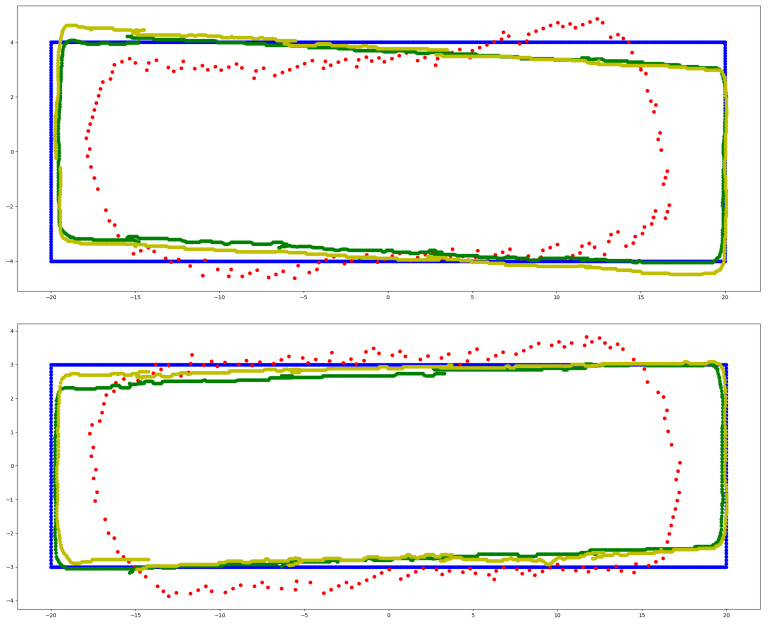}
\end{center}
   \caption{Visual comparison of different reconstruction methods. blue, red, yellow, green curves represent the ground truth of the crack profile, aligned-vision, passive-tactile and our method for crack reconstruction, respectively.}
\label{fig:reconstruction}
\end{figure}

As can be seen in Table \ref{tab:re3}, our approach shows a significant improvement of mean distance error from 0.55mm to 0.24mm compared to the aligned vision method. Two example reconstructed crack profiles are shown in Fig. \ref{fig:reconstruction}, which shows that reconstructed crack profiles with tactile data is much closer to the ground truth compared to the vision-based method. On the other side, compared to passive tactile perception, our proposed vision-guided tactile perception is more than 10 times faster than passive tactile perception in terms of the tactile data collection time without affecting the accuracy much. 


\begin{table}[h]
	\centering
		\caption{Reconstruction Accuracy}
		\label{tab:re3}
        \scalebox{1}{
		\begin{tabular}{c| c | c | c | c }
			\hline
			Method & MeanD(mm) & SD(mm) & MaxD(mm) & time(s)  \\
			\hline
			\hline
    		vision & 0.82 & 0.92 & 4.87 & \textbf{1} \\ 
    	    aligned-vision & 0.55 & 0.53 & 3.78 & \textbf{1} \\ 
    		passive-tactile  & \textbf{0.20} & 0.17& 0.99 & 400\\
    		active-tactile[ours]     & 0.24  &\textbf{0.16} & \textbf{0.82} & 35\\
    		\hline
		\end{tabular}}
\end{table}

\section{CONCLUSION AND FUTURE WORK}
In this paper we introduce a novel vision-guided active tactile perception for crack detection and reconstruction. The cooperation between those two modality addresses the false positives in visual detection results. The experiments show that our proposed method can improve the effectiveness and robustness of crack detection and reconstruction significantly, compared to when only vision is used. It has the potential to enable robots to inspect and repair of the concrete infrastructure. Future works to improve our method can also be considered, such as cracks detection in curved surfaces, the use of weakly supervised learning methods to segment cracks and addressing the localization uncertainty of the robot arm in active tactile exploration.

\bibliographystyle{IEEEtran}
\bibliography{egbib}

\end{document}